\documentclass{article}

\usepackage[final, nonatbib]{neurips_2021}
\usepackage{float}

\usepackage[utf8]{inputenc} 
\usepackage[T1]{fontenc}    
\usepackage{hyperref}       
\usepackage{url}            
\usepackage{booktabs}       
\usepackage{amsfonts}       
\usepackage{nicefrac}       
\usepackage{microtype}      
\usepackage{xcolor}         
\usepackage{graphicx}
\usepackage{subfig}

\usepackage[symbol*]{footmisc}
\setfnsymbol{wiley}

\usepackage[backend=bibtex, sorting=none]{biblatex}
\renewbibmacro{in:}{}

\bibliography{neurips_2021.bbl}

\usepackage{cleveref}
\title{Estimating Uncertainty For Vehicle Motion Prediction on Yandex Shifts Dataset}

\author{%
  Alexey Pustynnikov\thanks{equal contribution} \\
  VTB Bank\\
  \texttt{alexeypustynnikov@gmail.com} \\
  \And
  Dmitry Eremeev\footnotemark[1] \\
  Rosgosstrakh \\
  \texttt{dv.eremeev@yandex.ru} \\
}

\begin{document}

\maketitle

\begin{abstract}
  Motion prediction of surrounding agents is an important task in context of autonomous driving since it is closely related to driver's safety. Vehicle Motion Prediction (VMP) track of Shifts Challenge\footnote{\href{https://research.yandex.com/shifts/}{https://research.yandex.com/shifts/}} focuses on developing models which are robust to distributional
shift and able to measure uncertainty of their predictions.
In this work we present the approach that significantly improved provided benchmark and took 2nd place on the leaderboard.
\end{abstract}

\section{Introduction}
The task of multimodal trajectory prediction of road agents was thoroughly analyzed during past years and has led to the emergence of several groups of methods.
One common approach is to use Computer Vision (CV) models with rasterized scene images as inputs \cite{multimodaldcn}, \cite{hdmapsconvnets}.
However, powerful CV models have a large number of parameters, could be computationally intensive and quite slow on inference.
The disadvantages also include encoding a quite large amount of irrelevant to the underlying process information from the entire scene.
To overcome these drawbacks most state-of-the-art approaches lean on the graph structure \cite{VectorNet}, \cite{LaneGCN}.
Graph-based methods take into account only relevant to the driving patterns information and can provide better quality due
to the expressive power of graph neural networks \cite{GCN}, \cite{GAT}.
Several approaches further improve the results by increasing the complexity of trajectory decoders \cite{TNT}, \cite{mmTransformer}.

However, the problem of uncertainty estimation (UE) in the context of VMP has not been widely covered yet.
Previous works \cite{uncertaintydjuric}, \cite{aleatoric}, \cite{RIP} consider a limited number of methods and are
either disconnected from state-of-the-art approaches in VMP or recent advances
in UE field \cite{DUQ}, \cite{SNGP}, \cite{DUE} or both.
On the other hand, Bayesian Deep Learning field would benefit from benchmarking on large industrial datasets.

In this work we constructed and benchmarked the solution
that simultaneously tries to meet the needs of VMP task and uses recent advances in both VMP and UE fields.
In what follows, we describe the dataset and our solution.

\section{Approach}
In our work we adopt a single forward pass method for uncertainty estimation.
The resulting model has two parts: multimodal motion prediction component and external neural network \cite{single_det} to model the uncertainty.
The general setting is shown on the Figure 1.
Both components use graph representation of the scene which we describe in the following section and process them with graph neural networks (GNN).

\subsection{Problem Statement}
The Vehicle Motion prediction part of Shifts Dataset \cite{Shifts} contains 5 seconds of past and 5 seconds of future states for all agents in a scene
along with overall scene features. The goal of the challenge is to build a model that predicts $k \leq 5$ future trajectories $\tilde{y}_{i}^{k}$ in the horizon of $T=25$ timesteps
along with their confidences $\omega^{k}$
and overall scene uncertainty $U$ for each scene $\mathbf{x}_i$.

The dataset $\mathcal{D} = \{(\mathbf{x}_i, \mathbf{y}_i)\}_{i=1}^{N}$, where $\mathbf{x}_i$ are features of the scenes,
$\mathbf{y}_i \in \mathbb{R}^{T\times 2}$ are ground truth trajectories and $i$ is the scene index,
is divided into 3 parts: train part contains only scenes from Moscow with no precipitation, but
there are different locations and precipitations in development and evaluation parts in order to create data shift setting.

\subsection{Data Preprocessing}
We extract geometries for parts of the road lanes and crosswalk polygons from provided raw data
and use past observed trajectories of all agents in the scene as is.
We refer to these elementary geometries as polylines following \cite{VectorNet}.

Following this, we transform coordinates so that the origin is located at the target agent's last observed position,
and the vehicle is headed towards positive direction of x-axis.
We select points within squared bounding box of size $120$ meters centered at the origin in order to capture only relevant information.
After that we redistribute points of lane and crosswalk geometries to impose the constant distance of $d=2$ meters between adjacent points.
Finally, we create fully-connected subgraphs for each polyline and invert them.
In the end of all operations the whole scene is represented as a graph with fully-connected components which correspond to polylines.
An example of the processed scene is shown on the Figure 1.

\begin{figure}
  \centering
    \centering
    \raisebox{0.2\height}{\subfloat{{\includegraphics[scale=0.2]{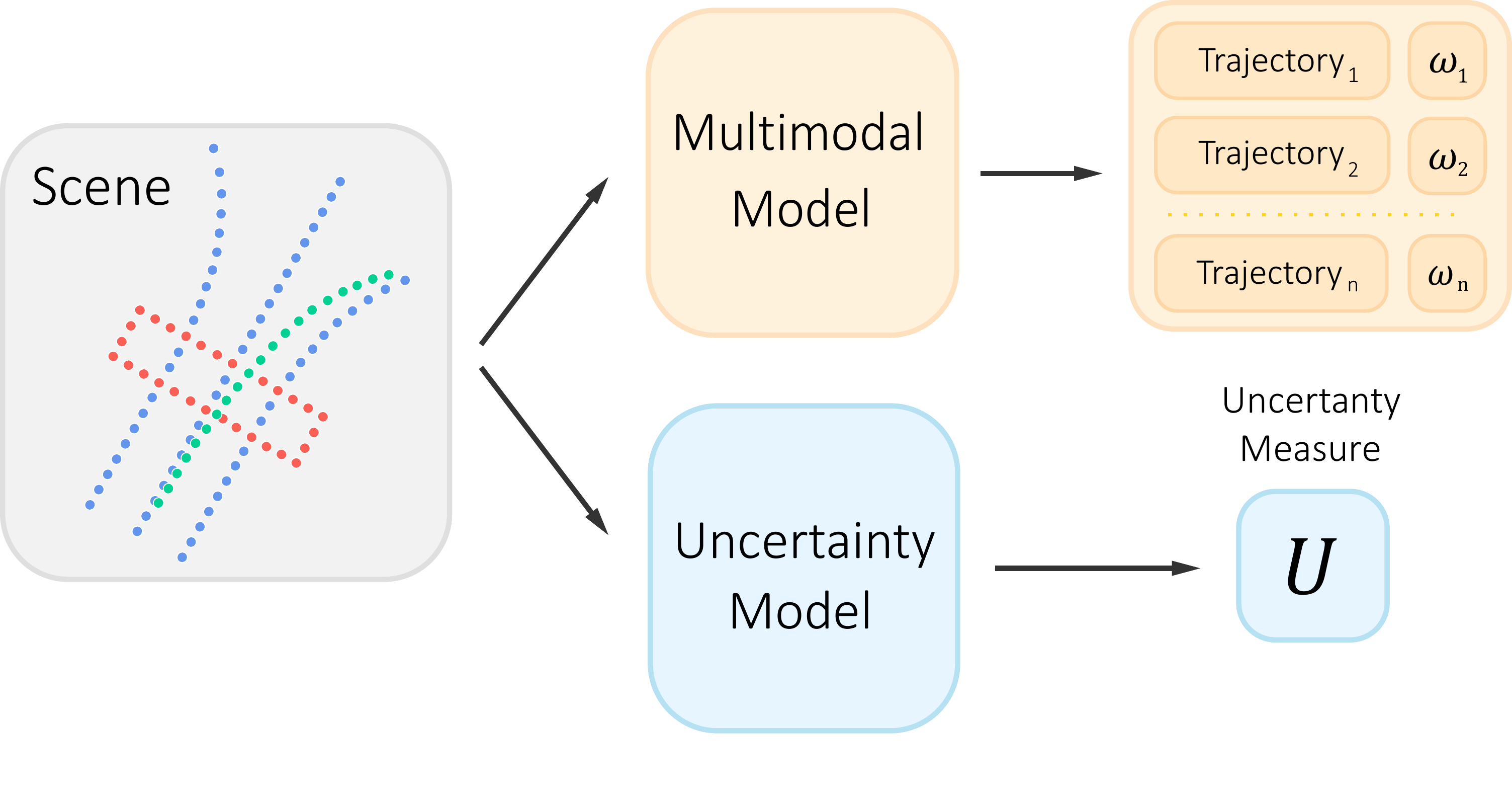} }}}%
    \qquad
    \subfloat{{\includegraphics[scale=0.125]{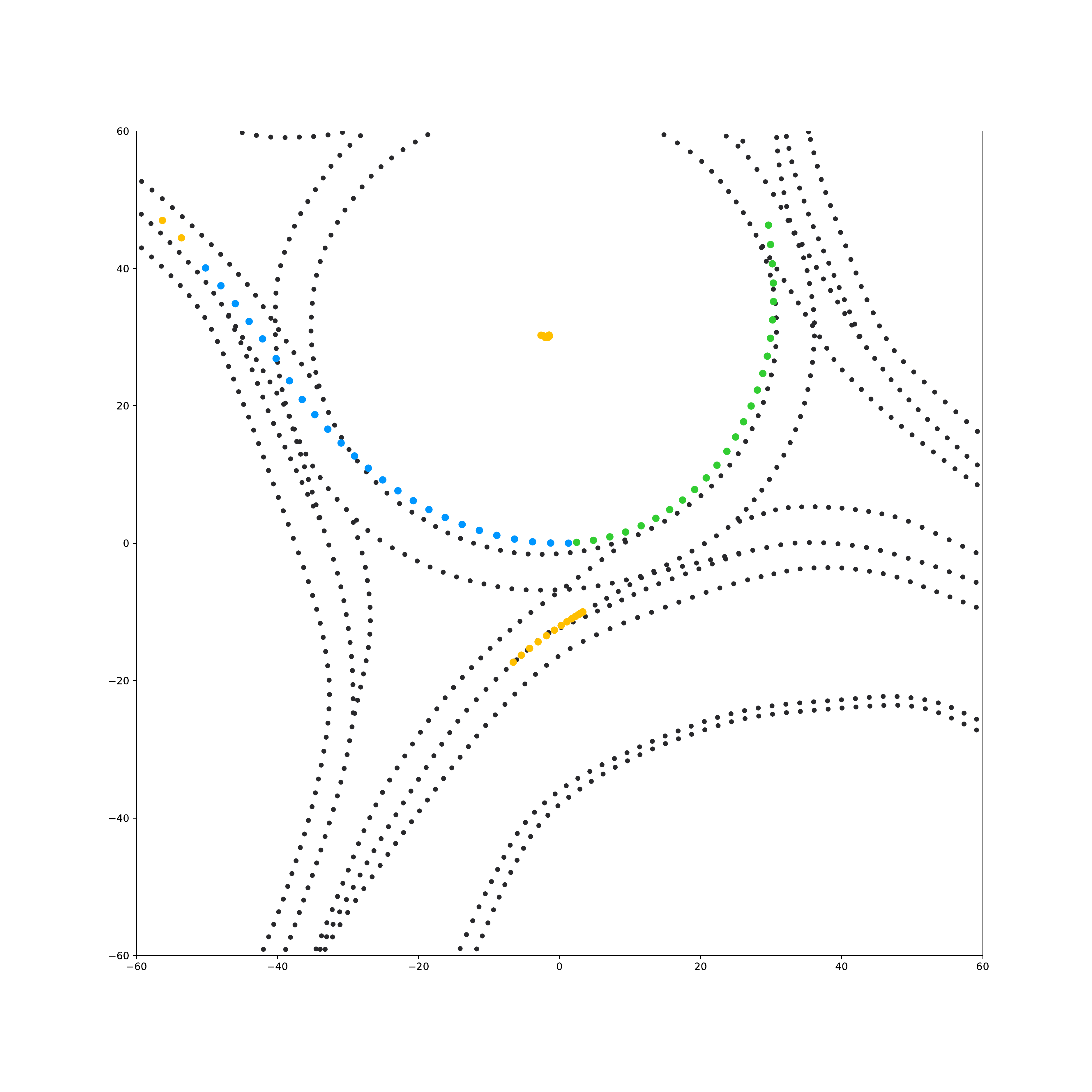} }}%
    \caption{(left) Structure of the composite model. (right) An example of the preprocessed scene. Black points represent lanes, blue and green - past and future target agent's trajectories respectively, orange depicts other agents in the scene.}
    \label{generalrender}%
\end{figure}

We construct the node feature-matrix of the form $X_i=(p_i^s, p_i^e, f_i)$, where $p_i^s \in \mathbb{R}^2 $, $p_i^e \in \mathbb{R}^2$ are
the coordinates of "start" and "end" adjacent points which formed a node after inversion,
$f_i \in \mathbb{R}^{9}$ are feature-vectors which we fill differently for various polyline types and $i$ is the node index.
Namely, we use maxspeed, lane priority and lane availability for lane polylines.
Lane availability status takes into account the traffic light state only at the latest available timestamp.
For agent polylines we fill in the following ones: timestamp, vectors of velocity and acceleration, yaw.
It is worth mentioning that in contrast to \cite{VectorNet} we do not use integer ids of polylines as features.

\subsection{Multimodal Model}
We adopt VectorNet \cite{VectorNet} architecture to obtain a hidden representation for each scene.
VectorNet is an hierarchical graph neural network that firstly builds vector representations for each polyline subgraph
and then propagates signal among all obtained representations.
The model is trained using corrected NLL of gaussian mixture as loss function: $\mathcal{L} = -\log \sum_{k}\omega^k \prod_{t=1}^{T} \mathcal{N}(y_t|\tilde{y}_{t}^k, I) - T\log(2\pi)$.

\begin{figure}
  \centering
   \includegraphics[scale=0.35]{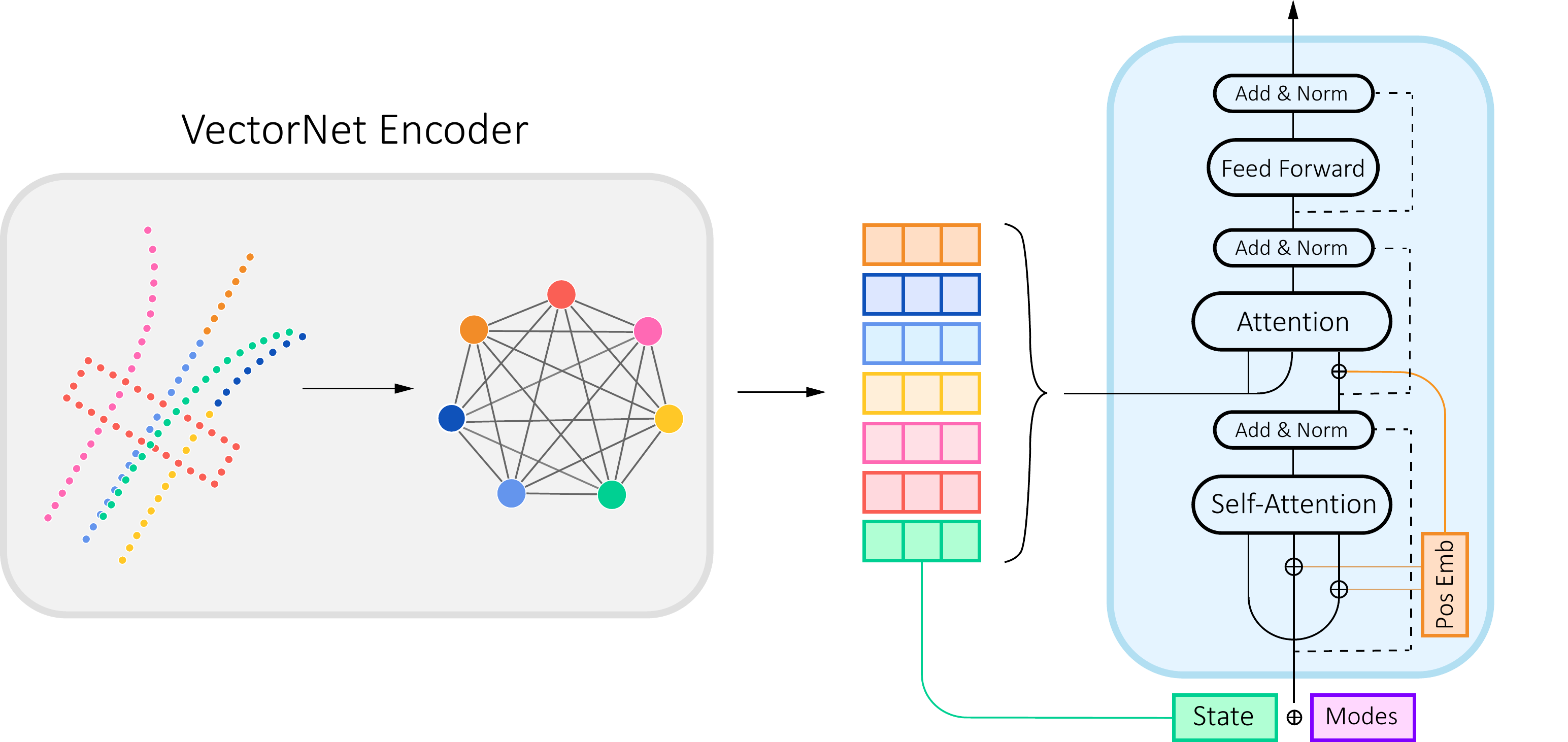}
    \caption{VectorNet-based Transformer.}%
    \label{fig:mmmodelstruct}%
\end{figure}

At the first stage we obtain embeddings for each node of fully connected subgraphs by applying Message Passing \cite{MessagePassing}:
\begin{equation}
h^{(t)}_i=\mathcal{U}^{(t)} \left(h^{(t-1)}_i, \Box_{j \in \mathcal{N}(i)} \mathcal{M}^{(t)} (h^{(t-1)}_i, h^{(t-1)}_j )\right),
\end{equation}
where $h_i^{(t)}$ - representation of node with index $i$ at layer $t$,
$\mathcal{N}(i)$ - neighbourhood of node $i$, which is formed by all other nodes in polyline in our case,
$\Box$ - aggregation function, $\mathcal{U}$ and $\mathcal{M}$ - update and message functions respectively.
We use $\max$ as aggregation function, MLP as message and concatenation as update.
After that we apply maxpooling on top of node embeddings in order to get vector representation $d_k$ for the whole polyline with index $k$.

At the second stage we construct fully-connected graph from all polyline representations $d$ and
apply graph transformer convolution \cite{graphtransfromer}:
\begin{equation}
 d^{(t)}_k = W_1 d^{(t-1)}_k +\sum_{l \in \mathcal{N}(k)} \alpha_{kl} W_2 d^{(t-1)}_{l}, \quad \alpha_{kl} = \textrm{softmax} \left((W_3 d^{(t-1)}_k)^{T} (W_4 d^{(t-1)}_l)/\sqrt{D} \right).
  \end{equation}
Here $W_m$ are weight matrices and $D$ is a hidden dimension.

Motivated by suggested in \cite{VectorNet} directions of future work and success of stacked-transformers
architecture \cite{mmTransformer} in motion prediction task,
we extend VectorNet model with Transformer-based decoder.
The whole architecture of the model is shown on the Figure 2.
We utilize all polyline embeddings from the previous step as keys and values.
Queries are formed by $k$ copies of target agent's embedding summed with learnable vectors for each trajectory mode.
Additionally, we add learnable positional embeddings to queries and keys in decoder self-attention and to queries
in encoder-decoder attention as it was suggested in \cite{mmTransformer}.

\subsection{Uncertainty Model}
We apply spectral-normalized Gaussian process \cite{SNGP} (SNGP) for uncertainty estimation.
Since obtaining good representations is crucial for this task, we pretrain plain VectorNet encoder with
shallow MLP layer on the same multimodal objective.
We exclude transformer decoder to ensure that information would not be shared between encoder and transformer decoder.
Obtained embeddings for development part of dataset are clearly separable by motion type (see Figure 3).

After that we train SNGP head on top of pretrained encoder using unimodal trajectory prediction task.
We insert single spectral normalized linear layer before gaussian-process output layer.
During the training process SNGP optimizes posterior probability $p(\tilde{y}_{i}|\mathcal{D})\sim p(\mathcal{D}|\tilde{y}_{i})p(\tilde{y}_{i})$
of target agent's trajectory  $\tilde{y}_{i} \in \mathbb{R}^{50}$ with Gaussian likelihood $p(\mathcal{D}|\tilde{y}_{i})$.
Random Fourier feature expansion \cite{RFF} with dimension $D_L$
approximates output as $\tilde{y}_{ik}=\Phi_{i}^{T}\beta_k\equiv\sqrt{2/D_{L}}\cos\left(-W_Ld_i+b_L\right)^{T}\beta_k$,
where $W_L \sim \mathcal{N}(0, I/l^2)$, $b_L \sim \mathcal{U}_{[0, 2\pi]}$
and $\beta$ has a normal prior.

SNGP further approximates $\beta_k$ posterior by using Laplace approximation $p(\beta_k|\mathcal{D}) \sim \mathcal{N}(\hat{\beta}_k, \hat{\Sigma}_k)$,
where $\hat{\beta}_k$ is MAP estimate obtained by gradient descent and the update rule
for $\hat{\Sigma}_k^{-1}$ with ridge factor $s$ and discount factor $m$ is written as follows: $\hat{\Sigma}_{k, 0}^{-1} = s\cdot I$,  $\hat{\Sigma}_{k, t}^{-1} = m \hat{\Sigma}_{k, t-1}^{-1} + (1-m) \sum_{i=1}\Phi_i\Phi_i^{T}$.
We estimate uncertainty as posterior variance for the scene: $U(\mathbf{x}_i) = \Phi_i^{T}\hat{\Sigma}\Phi_i$.

\begin{figure}
  \centering
    \centering
     \subfloat{{\includegraphics[scale=0.18]{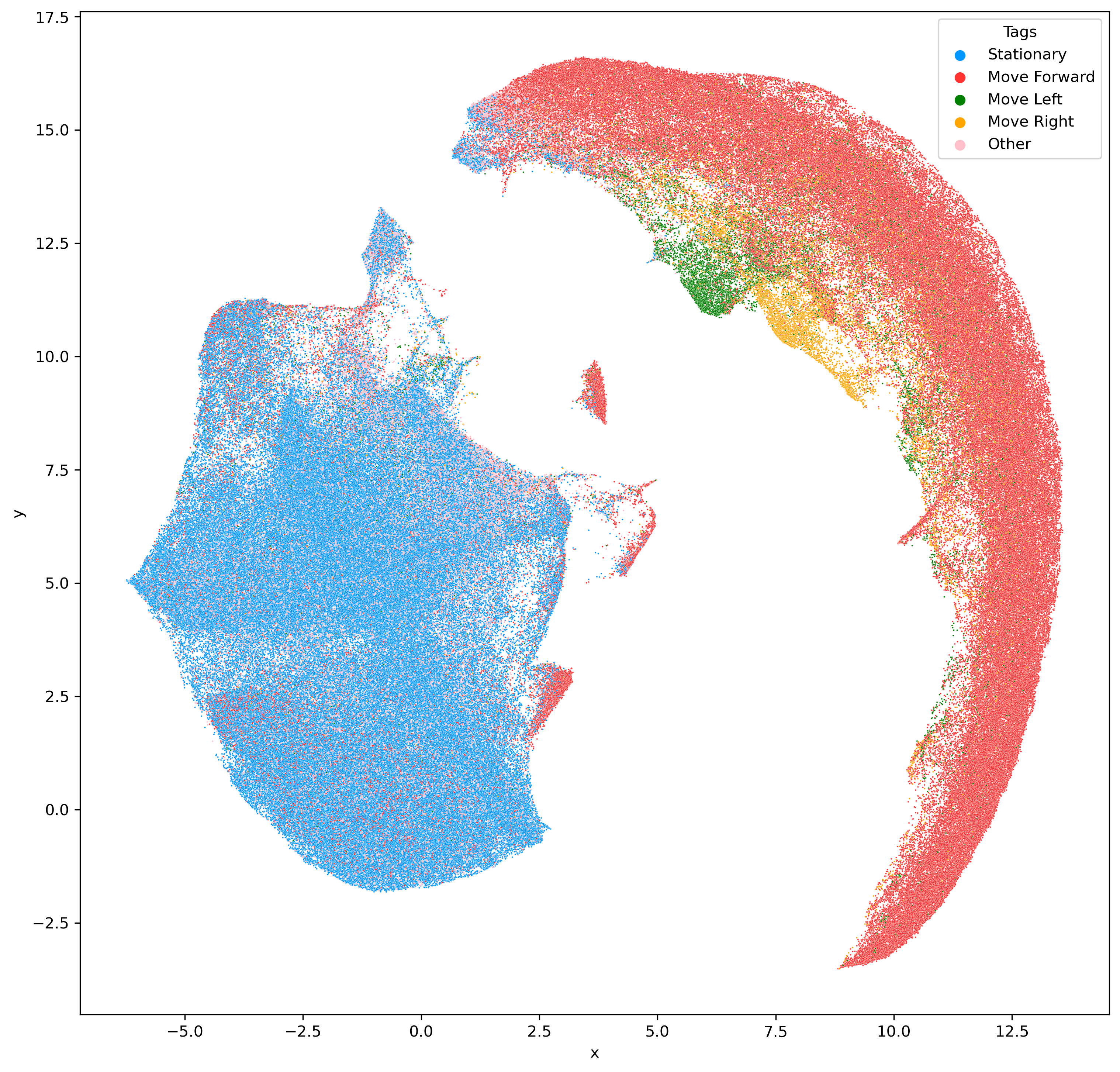} }}%
    \qquad
    \subfloat{{\includegraphics[scale=0.18]{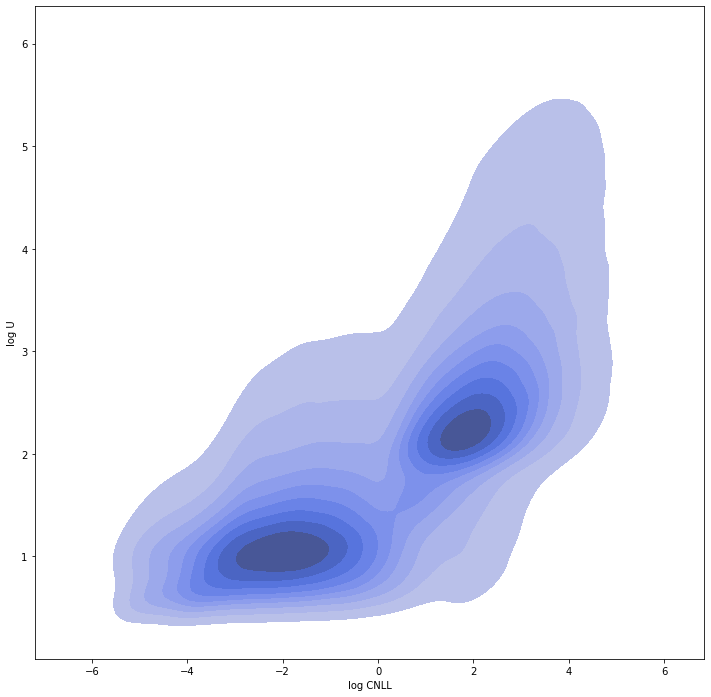} }}%
    \caption{(left) Visualization of embeddings obtained by UMAP \cite{UMAP}. Blue corresponds to stationary, red, green and orange correspond to moving forward, left and right respectively, pink stands for other. (right) Logarithm of uncertainty measure against
    logarithm of CNLL for development dataset.}%
    \label{fig:results}%
\end{figure}

\section{Experiments and Results}
The model was implemented using PyTorch \cite{PyTorch} and PyTorch Geometric \cite{PyG} and trained on single RTX 3090 GPU on $50\%$ of the data
with batch size of 32.

VectorNet encoder uses 3 layers of subgraph Message-Passing with hidden dimensions of 64 and transformer convolution layer with 2 attention heads of size 64.
The heads are averaged instead of concatenation.
Decoder has 2 transformer blocks with 4 heads and hidden dimension of 128.
Feed-Forward network has 2 linear layers with intermediate dimension of 256.
The inputs (state+mode embeddings) are not scaled before adding positional embeddings.

SNGP model utilizes pretrained encoder with the same parameters.
We set $D_L=2048$, $m=0.9999$, $s=0.1$ and $l^{-1}=0.05$.
Tuning these parameters could be quite tricky since SNGP is sensible to batch size and volume of data.

For all models the initial learning rate was set to $0.001$ and Adam optimizer was used.
For a multimodal model we used MultiStepLR scheduler with milestones $[6, 12]$
with a decay factor $0.3$ and the model was trained for a total of $20$ epochs.
We pretrained VectorNet encoder for $19$ epochs with decay every 5 epochs by a factor 0.3.
SNGP head was trained for a $5$ epochs.
We reset covariance matrix at the end of each epoch.

The results on the evaluation data are shown in the Table 1.
As it is shown on the Figure 3, there is an expected trend:
higher values of CNLL correlate with high values of uncertainty measure.
\begin{table}[H]
  \label{sample-table}
  \centering
  \begin{tabular}{ccccccc}
    \toprule
    Model         & R-AUC CNLL & CNLL     & minADE  & minFDE  & wADE    & wFDE\\
    \midrule
    VN SNGP       & $2.942$    & $17.037$ & $0.514$ & $1.001$ & $1.330$ & $3.167$ \\
    VNTransformer SNGP & $2.619$    & $15.599$ & $0.495$ & $0.936$ & $1.326$ & $3.158$ \\
    \bottomrule
  \end{tabular}
  \smallskip
  \caption{Results obtained by training on $50\%$ data in Shifts Challenge. VN stands for model with plain VectorNet, VNTransformer stands for model with Transformer decoder.
           Area under CNLL retention curve (R-AUC CNLL) \cite{Malininphd} was used for final ranking.}
\end{table}

\section{Conclusion}
  We considered a composite model for multimodal trajectory prediction and uncertainty estimation in vechicle motion prediction task.
  The resulting model shows a good performance in multimodal task and ability to detect out-of-distribution examples.
  \printbibliography

\end{document}